\documentclass[11pt]{article}

\usepackage[final]{acl}

\usepackage{times}
\usepackage{latexsym}
\usepackage{amsmath}
\usepackage{amssymb}
\usepackage{booktabs} 
\usepackage{graphicx}
\usepackage{algorithm}
\usepackage{algorithmic}

\usepackage[T1]{fontenc}
\usepackage[utf8]{inputenc}

\usepackage{microtype}
\usepackage{inconsolata}

\title{Graph-GRPO: Stabilizing Multi-Agent Topology Learning via Group Relative Policy Optimization}

\author{
  Yueyang Cang$^{1}$, Xiaoteng Zhang$^{1}$, Erlu Zhao$^{1}$, Zehua Ji$^{1}$,\\
  Yuhang Liu$^{1}$, Yuchen He$^{1}$, Zhiyuan Ning$^{1}$, Yijun Chen$^{1}$,\\
  Wenge Que$^{2,\star}$, Li Shi$^{1,\star}$\\
  \\
  $^{1}$Tsinghua University\\
  $^{2}$Donghua University\\
  $^{\star}$Corresponding author
}

\begin{document}
\maketitle

\begin{abstract}
Optimizing communication topology is fundamental to the efficiency and effectiveness of Large Language Model (LLM)-based Multi-Agent Systems (MAS). While recent approaches utilize reinforcement learning to dynamically construct task-specific graphs, they typically rely on single-sample policy gradients with absolute rewards (e.g., binary correctness). This paradigm suffers from severe gradient variance and the credit assignment problem: simple queries yield non-informative positive rewards for suboptimal structures, while difficult queries often result in failures that provide no learning signal. To address these challenges, we propose Graph-GRPO, a novel topology optimization framework that integrates Group Relative Policy Optimization. Instead of evaluating a single topology in isolation, Graph-GRPO samples a group of diverse communication graphs for each query and computes the advantage of specific edges based on their relative performance within the group. By normalizing rewards across the sampled group, our method effectively mitigates the noise derived from task difficulty variance and enables fine-grained credit assignment. Extensive experiments on reasoning and code generation benchmarks demonstrate that Graph-GRPO significantly outperforms state-of-the-art baselines, achieving superior training stability and identifying critical communication pathways previously obscured by reward noise.
\end{abstract}

\section{Introduction}

The rapid evolution of Large Language Models (LLMs) has catalyzed the development of Multi-Agent Systems (MAS), where collaborative agents demonstrate emergent capabilities in complex reasoning, coding, and decision-making tasks \cite{li2023camel, xi2025rise, hong2023metagpt, qian2023chatdev}. A growing number of studies suggest that the communication topology—the structural framework governing information exchange among agents—is a key determinant of system performance \cite{zhuge2024gptswarm, qian2025scaling, liu2023dynamic}. While early approaches relied on static, predefined structures such as chains, trees, or fully connected graphs \cite{wei2022chain, wu2023autogen, yao2024tree}, recent state-of-the-art methods like EIB-LEARNER \cite{shen2025understanding} have shifted towards dynamically generating task-specific topologies. EIB-LEARNER, for instance, provides a causal framework to balance ``error suppression'' and ``insight propagation'', demonstrating that adaptive connectivity is the key to robust collaboration \cite{zhang2025gdesigner, wang2024survey}.

\begin{figure*}[t]
\centering
\includegraphics[width=0.95\textwidth]{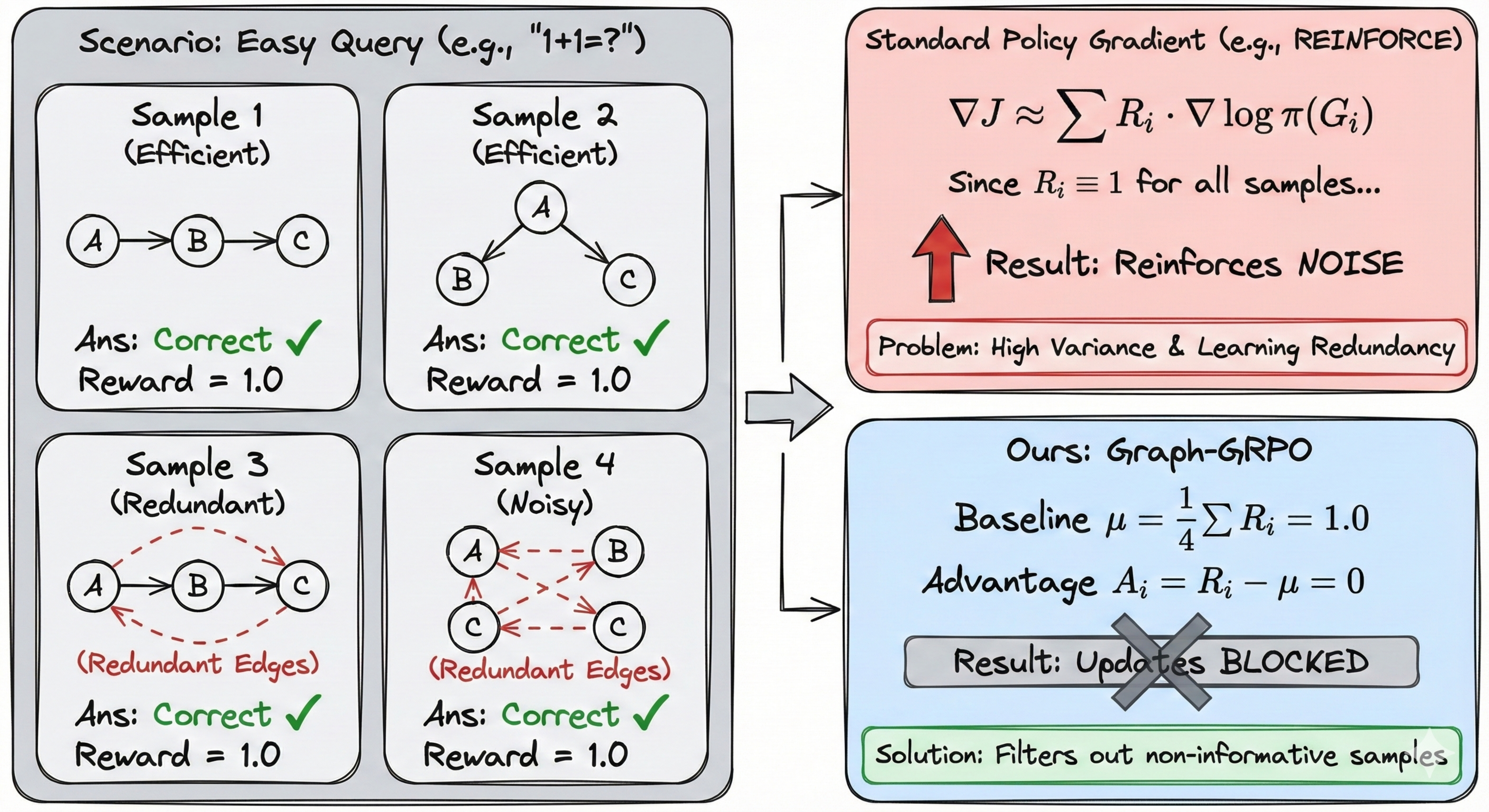} 
\caption{\textbf{Motivation Analysis: The Trap of Non-Informative Batches in Easy Queries.} 
The figure illustrates a scenario where a task is simple enough that diverse sampled topologies (Samples 1--4, ranging from efficient chains to dense structures with redundant edges) all yield correct answers and identical rewards ($R_k=1$).
(Top Right) Standard policy gradient methods like REINFORCE use raw rewards. Since $R_k \equiv 1$ across the entire group, the gradient estimation indiscriminately reinforces \textit{all} sampled edges, including noise and redundancies (e.g., extra edges in S3 \& S4), leading to suboptimal convergence.
(Bottom Right) Our proposed Graph-GRPO addresses this by incorporating a group baseline $\mu$. In such uniform-reward scenarios, $\mu$ equals individual rewards, resulting in near-zero advantage ($A_{ij} \approx 0$). This mechanism effectively blocks parameter updates from non-informative batches, preventing the model from learning redundant structures from noise.}
\label{fig:motivation_analysis}
\end{figure*}

Although topology modeling has advanced, the optimization paradigms for these discrete structures remain suboptimal. Most leading methods currently rely primarily on standard Reinforcement Learning (RL) techniques, such as the REINFORCE algorithm \cite{williams1992simple}, with single-sample estimation and absolute rewards (e.g., binary correctness) \cite{ouyang2022training}. This optimization strategy suffers from two fundamental limitations: 

\begin{enumerate}
    \item \textbf{High Gradient Variance:} The difficulty of queries in datasets is often uneven \cite{wang2023selfconsistency}. For simple queries, a wide range of suboptimal topologies may fortuitously yield correct answers (reward=1), introducing significant noise into the policy update. As illustrated in Figure \ref{fig:motivation_analysis}, standard methods indiscriminately reinforce these redundant edges. Conversely, for difficult queries, the system often fails regardless of the topology (reward=0), leading to vanishing gradients.
    \item \textbf{The Credit Assignment Problem:} When a topology succeeds, standard methods attribute the reward equally to all edges in the graph \cite{sutton2018reinforcement}. This coarse-grained feedback fails to distinguish which specific connections were causally responsible for the success and which were redundant, hindering the model's ability to learn precise structural patterns.
\end{enumerate}

To address these challenges, we propose Graph-GRPO (Graph-based Group Relative Policy Optimization), a novel framework that fundamentally stabilizes topology learning. Inspired by recent advances in LLM reasoning optimization \cite{shao2024deepseek, schulman2017proximal}, we shift the objective from maximizing absolute rewards to maximizing relative advantage within a sampled group. Specifically, for each query, Graph-GRPO samples a group of diverse communication topologies. Instead of evaluating each graph in isolation, we compute a baseline from the group's average performance and derive the advantage of each specific edge.

This group-based approach offers a dual benefit. First, it acts as a dynamic normalization mechanism: for simple tasks where the average performance is high, only topologies that perform better than average (e.g., more efficient) are reinforced, effectively filtering out ``easy-win'' noise. Second, it enables fine-grained credit assignment: edges that consistently appear in the higher-performing topologies within a group are assigned positive advantages, while those associated with failure are suppressed. By integrating this mechanism, Graph-GRPO allows the model to identify critical communication pathways that were previously obscured by reward noise.

In summary, our contributions are as follows:
\begin{itemize}
    \item We identify the limitations of absolute-reward optimization in MAS topology learning and propose Graph-GRPO, the first framework to apply Group Relative Policy Optimization to discrete structure search.
    \item We introduce a fine-grained edge scoring mechanism that solves the credit assignment problem by leveraging relative advantages across a group of sampled topologies.
    \item Extensive experiments on six benchmarks, including MMLU and HumanEval, demonstrate that Graph-GRPO significantly outperforms EIB-LEARNER, achieving superior stability and convergence efficiency.
\end{itemize}

\section{Related Work}

\subsection{LLM-based Multi-Agent Systems}
The paradigm of utilizing multiple Large Language Models (LLMs) to tackle complex tasks has garnered significant attention \cite{xi2025rise, wang2024survey}. Early frameworks such as CAMEL \cite{li2023camel} and AutoGen \cite{wu2023autogen} demonstrated that role-playing agents can collaboratively solve problems through dialogue. However, these initial systems typically operated on predefined, static communication structures, such as chain-of-thought sequences \cite{wei2022chain}, star topologies (centralized manager), or fully connected graphs \cite{hong2023metagpt, qian2023chatdev}. While effective for specific scenarios, static topologies lack the flexibility to adapt to the varying complexity of user queries, often leading to either redundant communication costs or insufficient information exchange \cite{liu2023dynamic, zhuge2024gptswarm}.

\subsection{Communication Topology Optimization}
To overcome the rigidity of static structures, recent research has focused on learning adaptive communication topologies. Approaches like AgentPrune \cite{zhang2025cut} and AgentDropout \cite{wang2025agentdropout} employ pruning techniques to remove redundant connections from a full graph. More advanced generative methods, such as G-Designer \cite{zhang2025gdesigner} and EIB-LEARNER \cite{shen2025understanding}, utilize Graph Neural Networks (GNNs) to construct task-specific topologies from scratch. EIB-LEARNER, in particular, introduced a causal perspective to balance error suppression and insight propagation. 

Despite these advances in topology modeling, the optimization strategy remains largely unchanged: these methods predominantly rely on standard policy gradient algorithms (e.g., REINFORCE) with absolute, binary rewards \cite{williams1992simple}. As noted in our analysis, this single-sample optimization paradigm suffers from high variance and poor credit assignment, especially when dealing with the diverse difficulty levels inherent in reasoning datasets. Our work builds upon the architectural strengths of EIB-LEARNER but fundamentally redesigns the optimization process to ensure stability and robustness.

\subsection{Reinforcement Learning for Reasoning}
Reinforcement Learning (RL) has become a foundational approach for aligning LLMs with human preferences and logical constraints \cite{ouyang2022training}. While Proximal Policy Optimization (PPO) \cite{schulman2017proximal} is widely used, its dependence on a value network (Critic) introduces significant memory overhead and training instability. Recently, Group Relative Policy Optimization (GRPO), introduced in DeepSeekMath \cite{shao2024deepseek}, has emerged as a powerful alternative. By eliminating the Critic and normalizing rewards within a sampled group, GRPO effectively reduces gradient variance for mathematical reasoning tasks.

However, existing applications of GRPO are largely confined to continuous text generation domains. To the best of our knowledge, our work is the first to adapt the group-relative mechanism to the domain of discrete structure search in multi-agent systems, addressing the unique challenges of edge-level credit assignment in graph topology learning.

\begin{figure*}[t]
\centering
\includegraphics[width=1.0\textwidth]{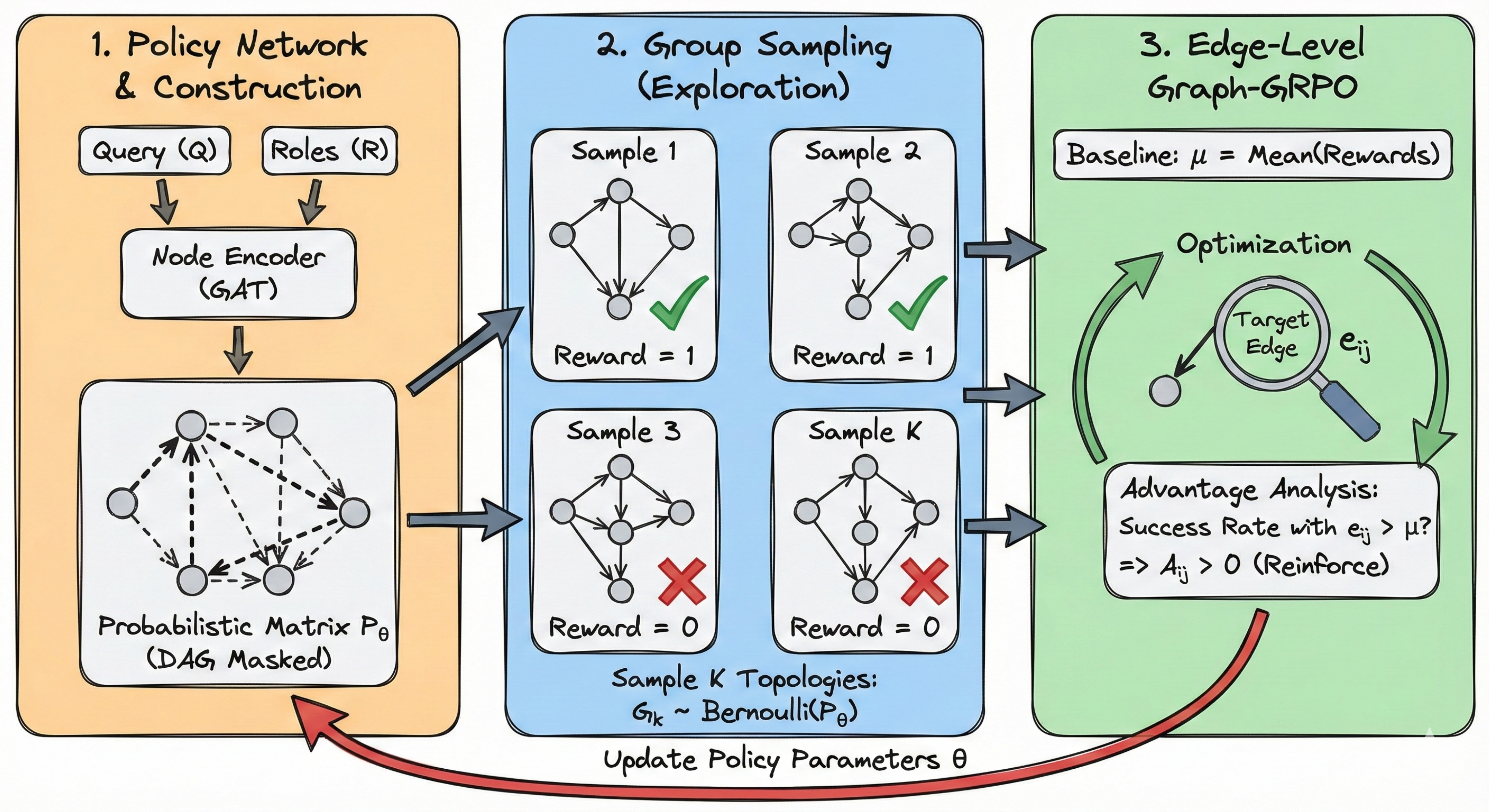} 
\caption{ \textbf{The overall framework ofGraph-GRPO}. 
\textbf{(1) Policy Network \& Construction:} The module encodes agent roles and the task query using a GAT-based encoder to generate a probabilistic connectivity matrix $P_\theta$, constrained by a DAG mask to ensure acyclic flow.
\textbf{(2) Group Sampling (Exploration):} Instead of a single estimation, we generate a group of $K$ diverse topologies via independent Bernoulli sampling. This exploration captures various structural patterns, where successful topologies receive positive rewards (Reward=1) and failures (e.g., disconnected graphs) receive zero.
\textbf{(3) Edge-Level Graph-GRPO:} The core optimization phase. We calculate a group baseline $\mu$ and estimate the specific advantage of each target edge $e_{ij}$. Edges that result in a success rate higher than the baseline ($A_{ij} > 0$) are reinforced, iteratively updating the policy parameters $\theta$.}
\label{fig:framework}
\end{figure*}

\begin{algorithm}[t]
\caption{Graph-GRPO Training Procedure}
\label{alg:graph_grpo}
\begin{algorithmic}[1]
\REQUIRE Training dataset $\mathcal{D}$, Group size $K$, Epochs $T$
\STATE Initialize policy network parameters $\theta$
\FOR{epoch $= 1$ to $T$}
    \FOR{each batch $(\mathcal{Q}, \text{Roles})$ in $\mathcal{D}$}
        \STATE Compute probability matrix $P_\theta$ via Eq. (2)
        \STATE \textbf{Sample Group:} Generate $K$ topologies $\{\mathcal{G}_1, \dots, \mathcal{G}_K\}$ via Bernoulli sampling (Eq. 3)
        \STATE \textbf{Evaluation:} Execute each $\mathcal{G}_k$ with LLM agents to get rewards $\{r_1, \dots, r_K\}$
        \FOR{each unique edge $(i,j)$ in the group}
            \STATE Calculate Conditional Success Rate $S_{ij}$ (Eq. 4)
        \ENDFOR
        \STATE Compute group stats $\mu_S, \sigma_S$ from all $\{S_{ij}\}$
        \STATE Compute Advantage $A_{ij}$ (Eq. 5)
        \STATE Update $\theta$ by minimizing $\mathcal{L}(\theta)$ (Eq. 6)
    \ENDFOR
\ENDFOR
\end{algorithmic}
\end{algorithm}

\section{Methodology}

In this section, we present the proposed Graph-GRPO framework. The overall architecture is depicted in Figure \ref{fig:framework}. We first outline the policy network architecture used to generate communication topologies, incorporating strict structural constraints to ensure logical progression. Then, we detail our core contribution: a group relative optimization mechanism that performs fine-grained credit assignment by estimating the marginal success rate of each edge, effectively eliminating the need for a value network (Critic).

\subsection{Policy Network Architecture}
We strictly followed the architectural design proposed in G-Designer \cite{zhang2025gdesigner} as our policy backbone. The framework utilizes a Graph Neural Network (GNN) to parameterize the communication topology and consists of two primary modules: a Node Encoder and a Structure Generator.

\paragraph{Node Representation.}
Given a task query $\mathcal{Q}$ and a set of agents $\mathcal{V}=\{v_1, \dots, v_N\}$, we first initialized the feature vector $x_i$ for each agent. Consistent with G-Designer, this was achieved by concatenating the agent's role description with the query content, followed by the pre-trained MiniLM encoder \cite{wang2020minilm}:
\begin{equation}
    x_i = \text{Encoder}(\text{Role}_i \oplus \mathcal{Q})
\end{equation}
where the encoder is fixed to the \texttt{all-MiniLM-L6-v2} checkpoint. This shared encoder ensures that agents with similar functional roles (e.g., two different ``Coder'' agents) perform similar topological behaviors, facilitating generalization.

\paragraph{Topology Generation with DAG Constraint.}
To capture the potential high-order dependencies between agents, we employed a multi-layer Graph Attention Network (GAT) \cite{velickovic2017graph}. We used a fully connected graph as the computational substrate for message passing. The GAT module updated agent embeddings by aggregating information from all other nodes, resulting in context-aware embeddings $H \in \mathbb{R}^{N \times D}$.

The probability of a directed connection from agent $v_j$ to $v_i$ was modeled via a bilinear inner product. Crucially, to ensure the reasoning process is acyclic and progressive, we applied a Directed Acyclic Graph (DAG) mask prior to activation. This inductive bias enforced $(P_\theta)_{ij} = 0$ for all $j \le i$, constraining information to flow strictly from earlier agents to later ones (typically converging towards the final agent $v_N$). The valid connection probabilities are computed as:
\begin{equation}
    (P_\theta)_{ij} = \begin{cases} 
    \sigma(h_i W h_j^T) & \text{if } j < i \\
    0 & \text{otherwise}
    \end{cases}
\end{equation}
where $W \in \mathbb{R}^{D \times D}$ is a learnable weight matrix modeling the affinity between roles, and $\sigma(\cdot)$ is the sigmoid function. This continuous probability matrix $(P_\theta)$ serves as the basis for both stochastic sampling during training and deterministic thresholding during inference.

\subsection{Graph-GRPO Optimization}
Standard policy gradient methods, such as REINFORCE \cite{williams1992simple}, assign a uniform reward to all edges in a graph. This creates a coarse-grained feedback loop where redundant edges in a successful graph are falsely reinforced, while critical edges in a failed graph are unfairly penalized. Inspired by Group Relative Policy Optimization (GRPO) \cite{shao2024deepseek}, we propose an Edge-Level Graph-GRPO strategy. Unlike PPO \cite{schulman2017proximal}, our method does not require a separate Critic network, reducing memory overhead and training instability.

\subsubsection{Group Sampling via Monte Carlo Approximation}
For each query $\mathcal{Q}$, we approximated the gradient expectation by sampling a group of $K$ distinct topologies $\{\mathcal{G}_1, \dots, \mathcal{G}_K\}$ from the current policy $\pi_\theta$. 
To ensure structural diversity and enable the exploration of various reasoning paths, we employed a stochastic sampling strategy. 
Specifically, the binary existence of an edge in the $k$-th sampled topology is determined by independent Bernoulli sampling parameterized by the predicted probabilities:
\begin{equation}
    \mathbb{I}((i,j) \in \mathcal{G}_k) \sim \text{Bernoulli}((P_\theta)_{ij})
\end{equation}
This probabilistic process transforms the continuous probability matrix into discrete graph structures. Crucially, this stochasticity allows the model to explore different connectivity patterns (e.g., sparse chains vs. dense trees) within the same group, constructing a robust local baseline from the group's own statistics for the subsequent advantage estimation.

\subsubsection{Marginal Success Rate Estimation}
To quantify the contribution of specific connections, we define an edge-specific score $S_{ij}$. The core intuition is \textit{counterfactual reasoning}: if an edge $e_{ij}$ is truly beneficial, its presence should be positively correlated with task success within the group. We calculate $S_{ij}$ as the conditional success rate:
\begin{equation}
    S_{ij} = \frac{\sum_{k=1}^K \big( \mathbb{I}((i,j) \in \mathcal{G}_k) \cdot r_k \big)}{\sum_{k=1}^K \mathbb{I}((i,j) \in \mathcal{G}_k) + \epsilon}
\end{equation}
where $r_k \in \{0, 1\}$ is the binary reward of the $k$-th topology, and $\epsilon$ is a small constant for numerical stability. 
The numerator represents the number of correct trials where edge $e_{ij}$ was active, while the denominator represents the total number of trials containing $e_{ij}$. Consequently, $S_{ij} \in [0, 1]$ estimates the empirical probability $P(\text{Success} | e_{ij} \in \mathcal{G})$. This mechanism effectively distinguishes critical pathways (high $S_{ij}$) from noise edges (random $S_{ij} \approx \text{Group Average}$).

\subsubsection{Relative Advantage and Objective}
To mitigate the variance caused by varying task difficulties (e.g., simple tasks yield high success rates for all edges), we applied the GRPO principle to normalize these scores. The advantage $A_{ij}$ is computed as:
\begin{equation}
    A_{ij} = \frac{S_{ij} - \mu_{S}}{\sigma_{S} + \epsilon}
\end{equation}
where $\mu_{S}$ and $\sigma_{S}$ are the mean and standard deviation of the scores $\{S_{ij}\}$ computed across \textit{all active edges} in the current group.
This normalization ensures that only edges contributing \textit{more than average} to the success rate receive positive reinforcement ($A_{ij} > 0$), while less effective edges are suppressed ($A_{ij} < 0$).

Following the standard formulation of GRPO \cite{shao2024deepseek}, we incorporated a KL-divergence term to constrain the policy update, preventing the model from deviating excessively from the initial distribution. The final loss function is defined as:
\begin{equation}
\begin{split}
    \mathcal{L}(\theta) = \frac{1}{|\mathcal{E}_{batch}|} \sum_{(i,j) \in \mathcal{E}_{batch}} \bigg( & - A_{ij} \log \pi_\theta(e_{ij} | \mathcal{Q}) \\
    & + \beta D_{KL}(\pi_\theta || \pi_{ref}) \bigg)
\end{split}
\end{equation}
where $\pi_{ref}$ represents the reference policy (initialized with the supervised fine-tuned parameters and frozen during RL training), and $\beta$ is the coefficient controlling the KL penalty strength. $D_{KL}$ denotes the Kullback-Leibler divergence between the current policy $\pi_\theta$ and the reference policy $\pi_{ref}$ for the specific edge distribution. This regularization ensures training stability and prevents reward hacking.

The complete training procedure is summarized in Algorithm \ref{alg:graph_grpo}.

\subsection{Inference Mechanism}
During the inference phase, we adopt a deterministic strategy to ensure reproducibility and stability. Given a test query $\mathcal{Q}$, we first computed the probability matrix $P_\theta$ using the trained policy network. To derive the final discrete topology $\mathcal{G}^*$, we applied a hard thresholding operation:
\begin{equation}
    \mathbb{I}((i,j) \in \mathcal{G}^*) = \begin{cases} 
    1 & \text{if } (P_\theta)_{ij} > \tau \\
    0 & \text{otherwise}
    \end{cases}
\end{equation}
where $\tau$ is a hyperparameter set to $0.5$. This mechanism effectively filters out low-confidence connections, resulting in a sparse, task-specific communication structure that minimizes redundancy while preserving critical reasoning pathways.

\begin{table*}[t]
\centering
\resizebox{\textwidth}{!}{
\begin{tabular}{l|cccccc|c}
\toprule
\textbf{Method} & \textbf{MMLU} & \textbf{GSM8K} & \textbf{AQuA} & \textbf{MultiArith} & \textbf{SVAMP} & \textbf{HumanEval} & \textbf{Avg.} \\
\midrule
Vanilla & 80.39 & 82.30 & 71.06 & 93.09 & 86.55 & 71.39 & 80.80 \\
\midrule
CoT & 81.69 & 86.50 & 73.58 & 93.25 & 87.36 & 74.67 & 82.84 \\
SC (CoT) & 83.66 & 81.60 & 75.63 & 94.12 & 88.59 & 79.83 & 83.91 \\
\midrule
Chain & 83.01 & 88.30 & 74.05 & 93.27 & 87.17 & 81.37 & 84.53 \\
Tree & 81.04 & 85.20 & 71.23 & 93.68 & 88.91 & 80.53 & 83.43 \\
Complete & 82.35 & 80.10 & 72.95 & 94.53 & 84.01 & 79.03 & 82.16 \\
Random & 84.31 & 86.90 & 76.48 & 94.08 & 87.54 & 82.66 & 85.33 \\
LLM-Debate & 84.96 & 91.40 & 77.65 & 96.36 & 90.11 & 84.70 & 87.53 \\
\midrule
AgentPrune & 85.07 & 91.10 & 80.51 & 94.65 & 90.58 & 86.75 & 88.09 \\
AgentDropout & 85.62 & 91.70 & 80.94 & 95.60 & 91.04 & 85.98 & 88.48 \\
G-designer & 86.92 & 93.80 & 81.60 & 96.50 & 93.10 & 88.33 & 90.04 \\
EIB-LEARNER & \underline{88.90} & \underline{95.20} & \underline{83.49} & \underline{96.83} & \underline{94.70} & \underline{89.15} & \underline{91.38} \\
\midrule
\textbf{Graph-GRPO} & \textbf{90.12} & \textbf{96.10} & \textbf{84.21} & \textbf{97.07} & \textbf{96.01} & \textbf{91.25} & \textbf{92.45} \\
\bottomrule
\end{tabular}
}
\caption{Performance comparison (\%) on six benchmarks. The best results are highlighted in \textbf{bold}, and the second best are \underline{underlined}. Baseline results are retrieved from \cite{shen2025understanding}.}
\label{tab:main_results}
\end{table*}

\begin{table}[h]
\centering
\resizebox{\columnwidth}{!}{
\begin{tabular}{l|ccc|c}
\toprule
\textbf{Method} & \textbf{MMLU} & \textbf{GSM8K} & \textbf{HumanEval} & \textbf{Avg.} \\
\midrule
\textbf{Graph-GRPO} & \textbf{90.12} & \textbf{96.10} & \textbf{91.25} & \textbf{92.49} \\
Graph-Level GRPO & 88.54 & 94.40 & 89.07 & 90.67 \\
\midrule
$\Delta$ & -1.58 & -1.70 & -2.18 & -1.82 \\
\bottomrule
\end{tabular}
}
\caption{Ablation study on optimization granularity: Edge-Level vs. Graph-Level.}
\label{tab:ablation}
\end{table}

\begin{figure*}[t]
\centering
\includegraphics[width=0.9\textwidth]{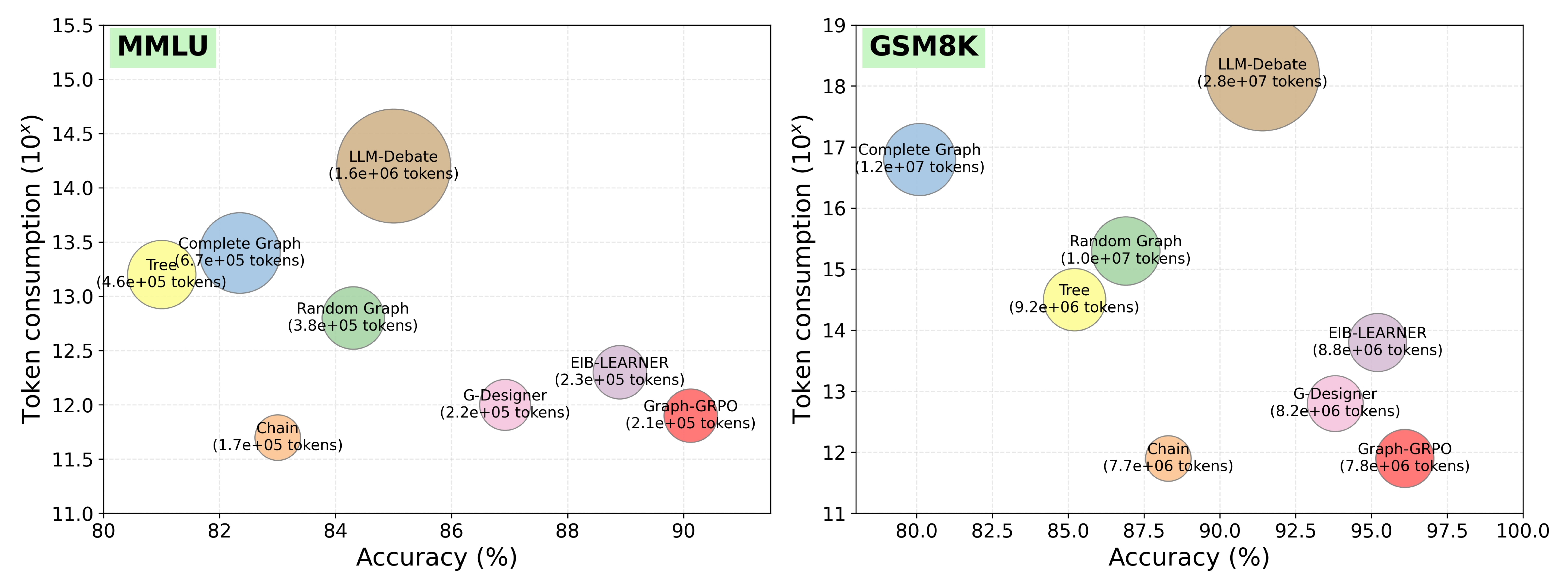} 
\caption{Token efficiency analysis on MMLU and GSM8K benchmarks. The bubble size represents the relative token consumption. \textbf{Graph-GRPO (Red)} achieves the highest accuracy (positioned furthest to the right) while maintaining a low token cost comparable to \textbf{EIB-LEARNER (Purple)} and \textbf{G-Designer (Pink)}. Our method effectively suppresses redundant edges without explicit pruning constraints, achieving a superior performance-efficiency trade-off compared to complete graphs (Blue) and debate-based baselines (Brown).}
\label{fig:token_efficiency}
\end{figure*}

\section{Experiments}
\subsection{Experimental Setup}

\paragraph{Datasets.}
Following the standard protocol in EIB-LEARNER \cite{shen2025understanding}, we evaluated our method on six benchmarks across three domains. For general reasoning, we used MMLU \cite{hendrycks2020measuring} to assess multi-task knowledge. In the mathematical domain, we employed four widely-used datasets: GSM8K \cite{cobbe2021training}, MultiArith \cite{roy2015solving}, SVAMP \cite{patel2021nlp}, and AQUA \cite{ling2017program}. Additionally, we used HumanEval \cite{chen2021evaluating} to evaluate code generation capabilities.

\paragraph{Baselines.}
We compared Graph-GRPO against three categories of baselines: (1) Single-Agent Methods, including Chain-of-Thought (CoT) \cite{wei2022chain} and Self-Consistency (SC) \cite{wang2023selfconsistency}; (2) Fixed Topologies, covering standard structures such as Chain, Tree, Complete Graph, and LLM-Debate \cite{du2023improving}; and (3) Topology Optimization Methods, which serve as our primary competitors, including AgentPrune \cite{zhang2025cut}, AgentDropout \cite{wang2025agentdropout}, G-Designer \cite{zhang2025gdesigner}, and EIB-LEARNER \cite{shen2025understanding}.

\paragraph{Implementation Details.}
We employed GPT-3.5-Turbo as the backbone LLM. The policy network utilized the \texttt{all-MiniLM-L6-v2} encoder and a 3-layer GAT, strictly aligned with G-Designer. The agent number $N$ was set to 6 for MMLU, 5 for HumanEval, and 4 for mathematical tasks. During training, we set the group sampling size $K=16$ and maximize communication rounds to 3. Optimization was performed via Adam with a learning rate of $1e-4$ on NVIDIA A100 GPUs.

\subsection{Main Results}

Graph-GRPO achieves state-of-the-art performance on all six benchmarks, demonstrating superior adaptability across diverse domains. As presented in Table \ref{tab:main_results}, Graph-GRPO attains the highest average accuracy of \textbf{92.45\%}, establishing a new benchmark for topology learning.

\paragraph{Comparison with Fixed Structures.}
Traditional static topologies (Chain, Tree, Complete) struggle to adapt to varying query complexities, capping their average performance at roughly 84\%. Notably, while the Complete Graph allows for maximum information flow, it suffers from a lower accuracy (82.16\%) compared to simpler structures. This counter-intuitive result highlights the detrimental effect of "information overload" and noise propagation in uncontrolled communication, validating the necessity of topology pruning.

\paragraph{Comparison with SOTA Optimization Methods.}
Compared to previous dynamic topology methods, Graph-GRPO shows distinct advantages. While EIB-LEARNER represents a strong baseline (91.38\%), its reliance on standard policy gradients limits its potential on harder tasks. Graph-GRPO outperforms EIB-LEARNER by a significant margin on complex reasoning benchmarks, such as \textbf{+0.9\% on GSM8K} and \textbf{+2.1\% on HumanEval}. This indicates that as task difficulty increases, the stability provided by our group-relative objective becomes increasingly critical. The overall improvement of \textbf{1.07\%} over the previous state-of-the-art confirms that our fine-grained credit assignment strategy successfully uncovers more effective reasoning pathways that were previously obscured by optimization noise.

\subsection{Ablation Study}
To investigate the source of our performance gains, we conducted a rigorous ablation study comparing our Edge-Level Graph-GRPO with a coarse-grained Graph-Level variant.

\paragraph{Graph-Level GRPO.}
In this variant, we assign the same advantage score to all edges within a sampled topology based on the graph's final result. This simulates a scenario where the ``credit assignment problem'' is not addressed.

\paragraph{Analysis of Degradation.}
Table \ref{tab:ablation} reveals a consistent performance degradation across all tasks when switching to Graph-Level optimization, with an average drop of \textbf{1.82\%}. The decline is particularly pronounced in HumanEval (-2.18\%), a task requiring precise logic chains.
This degradation substantiates our hypothesis: Graph-level rewards introduce severe structural noise. In a successful topology, not all edges are beneficial; some may be redundant or irrelevant. By rewarding the entire graph uniformly, the Graph-Level baseline reinforces these ``freeloader'' edges. Over time, this leads to denser, noisier graphs that hinder reasoning.
In contrast, Graph-GRPO's edge-level estimation acts as a soft filter. By aggregating statistics over $K$ samples, it isolates the marginal contribution of each edge, ensuring that only connections causally linked to success are reinforced. This fine-grained granularity is the cornerstone of our framework's robustness.

\subsection{Token Efficiency}
Beyond accuracy, economic efficiency is paramount for scalable MAS. We analyze the token consumption of Graph-GRPO relative to its performance in Figure \ref{fig:token_efficiency}.


\paragraph{Pareto Superiority.}
As illustrated in Figure \ref{fig:token_efficiency}, Graph-GRPO occupies the Pareto-optimal frontier (bottom-right corner), offering the best trade-off between cost and accuracy. Traditional methods like LLM-Debate or Complete Graphs incur prohibitive costs (high vertical position) due to quadratic message passing complexity ($O(N^2)$).
Crucially, Graph-GRPO achieves a token usage level comparable to explicit pruning methods like AgentPrune, yet delivers significantly higher accuracy. This implies that our method naturally converges to sparse but semantic topologies. By accurately identifying and penalizing non-informative edges during training, Graph-GRPO reduces the ``cognitive load'' on the system. It demonstrates that the key to efficiency is not merely cutting edges randomly, but preserving the high-value information pathways while eliminating noise, thereby maximizing the ``Signal-to-Token Ratio''.

\section{Conclusion}

In this work, we introduce \textbf{Graph-GRPO}, a novel framework that stabilizes multi-agent topology learning by fundamentally shifting the optimization paradigm from absolute rewards to group-relative advantage. By implementing a fine-grained edge-level score estimation strategy, our method successfully decouples structural optimization from the noise of task difficulty, effectively resolving the long-standing credit assignment problem in discrete topology search. Extensive evaluations across six reasoning and coding benchmarks demonstrate that Graph-GRPO not only establishes a new state-of-the-art but also naturally converges to sparse, semantic-rich structures, achieving a Pareto-optimal trade-off between decision accuracy and token efficiency. We believe this critic-free, variance-reduced paradigm paves the way for scalable, self-organizing agent swarms, with future work poised to extend this mechanism to larger-scale heterogeneous systems and open-ended, dynamic environments.

\section{Limitations}
While Graph-GRPO demonstrates strong performance, we acknowledge two main limitations. First, regarding scalability, our policy network relies on a GAT backbone with $\mathcal{O}(N^2)$ complexity. While efficient for typical reasoning groups ($N \le 6$), applying it to massive swarms (e.g., $N > 100$) may encounter computational bottlenecks, necessitating hierarchical or sparse generation strategies. Second, regarding dynamic adaptability, our framework generates a single static topology for each query. For complex, multi-turn dialogues where optimal communication structures might shift across turns, a finer-grained, turn-level topology adjustment mechanism would be more ideal.




\bibliography{custom}

@article{shen2025understanding,
  title={Understanding the Information Propagation Effects of Communication Topologies in LLM-based Multi-Agent Systems},
  author={Shen, Xu and Liu, Yixin and Dai, Yiwei and Wang, Yili and Miao, Rui and Tan, Yue and Pan, Shirui and Wang, Xin},
  journal={arXiv preprint arXiv:2505.23352},
  year={2025},
  note={The EIB-LEARNER paper}
}

@misc{shao2024deepseek,
  title={DeepSeekMath: Pushing the Limits of Mathematical Reasoning in Open Language Models},
  author={Shao, Zhihong and Wang, Peiyi and Zhu, Qihao and Xu, Runxin and Song, Junxiao and Xiao, Mingchuan and Yang, Y and others},
  year={2024},
  eprint={2402.03300},
  archivePrefix={arXiv},
  primaryClass={cs.CL},
  note={Origin of Group Relative Policy Optimization (GRPO)}
}

@article{schulman2017proximal,
  title={Proximal Policy Optimization Algorithms},
  author={Schulman, John and Wolski, Filip and Dhariwal, Prafulla and Radford, Alec and Klimov, Oleg},
  journal={arXiv preprint arXiv:1707.06347},
  year={2017}
}

@article{williams1992simple,
  title={Simple statistical gradient-following algorithms for connectionist reinforcement learning},
  author={Williams, Ronald J},
  journal={Machine learning},
  volume={8},
  number={3},
  pages={229--256},
  year={1992},
  publisher={Springer}
}

@inproceedings{ouyang2022training,
  title={Training language models to follow instructions with human feedback},
  author={Ouyang, Long and Wu, Jeffrey and Jiang, Xu and Almeida, Diogo and Wainwright, Carroll and Mishkin, Pamela and Zhang, Chong and others},
  booktitle={Advances in Neural Information Processing Systems (NeurIPS)},
  volume={35},
  pages={27730--27744},
  year={2022}
}

@inproceedings{zhuge2024gptswarm,
  title={GPTSwarm: Language Agents as Optimizable Graphs},
  author={Zhuge, Mingchen and Wang, Wenyi and Kirsch, Louis and Faccio, Francesco and Khizbullin, Dmitrii and Schmidhuber, J{\"u}rgen},
  booktitle={Proceedings of the 41st International Conference on Machine Learning (ICML)},
  year={2024}
}

@inproceedings{zhang2025gdesigner,
  title={G-designer: Architecting Multi-agent Communication Topologies via Graph Neural Networks},
  author={Zhang, Guibin and Yue, Yanwei and Sun, Xiangguo and Wan, Guancheng and Yu, Miao and Fang, Junfeng and Wang, Kun and Chen, Tianlong and Cheng, Dawei},
  booktitle={Proceedings of the 42nd International Conference on Machine Learning (ICML)},
  year={2025}
}

@inproceedings{liu2023dynamic,
  title={Dynamic LLM-Agent Network: An LLM-agent Collaboration Framework with Agent-Team Optimization},
  author={Liu, Zeyu and Yao, Huimo and Zhang, Chaowei and Yang, Zihuai and Tang, Jiakai and Yuan, Ye and Chen, Xu and Lin, Yankai and Sun, Maosong},
  booktitle={International Conference on Learning Representations (ICLR)},
  year={2024}
}

@inproceedings{li2023camel,
  title={CAMEL: Communicative Agents for "Mind" Exploration of Large Language Model Society},
  author={Li, Guohao and Hammoud, Hasan and Itani, Hani and Khizbullin, Dmitrii and Ghanem, Bernard},
  booktitle={Advances in Neural Information Processing Systems (NeurIPS)},
  volume={36},
  pages={51991--52008},
  year={2023}
}

@inproceedings{hong2023metagpt,
  title={MetaGPT: Meta Programming for A Multi-Agent Collaborative Framework},
  author={Hong, Sirui and Zheng, Xiawu and Chen, Jonathan and Cheng, Yuheng and Zhang, Ceyao and Wang, Ziyang and Yau, Steven Ka Shing and Lin, Zijuan and Zhou, Liyang and others},
  booktitle={International Conference on Learning Representations (ICLR)},
  year={2024}
}

@inproceedings{qian2023chatdev,
  title={ChatDev: Communicative Agents for Software Development},
  author={Qian, Chen and Liu, Wei and Liu, Hong and Chen, Nuo and Dang, Yufan and Li, Guohao and Yang, Cheng and Chen, Weize and Su, Yusheng and Liu, Zhiyuan and others},
  booktitle={Proceedings of the 62nd Annual Meeting of the Association for Computational Linguistics (ACL)},
  year={2024}
}

@inproceedings{qian2025scaling,
  title={Scaling Large-Language-Model-based Multi-Agent Collaboration},
  author={Qian, Chen and Xie, Zihao and Wang, Yifei and Liu, Wei and Dang, Yufan and Du, Zhuoyun and Chen, Weize and Yang, Cheng and Liu, Zhiyuan and Sun, Maosong},
  booktitle={International Conference on Learning Representations (ICLR)},
  year={2025}
}

@article{wu2023autogen,
  title={AutoGen: Enabling Next-Gen LLM Applications via Multi-Agent Conversation},
  author={Wu, Qingyun and Bansal, Gagan and Zhang, Jieyu and Wu, Yiran and Li, Beibin and Peng, Erkang and Wang, Xiubo and Zhang, Shaokun},
  journal={arXiv preprint arXiv:2308.08155},
  year={2023}
}

@article{xi2025rise,
  title={The Rise and Potential of Large Language Model Based Agents: A Survey},
  author={Xi, Zhiheng and Chen, Wenxiang and Guo, Xin and He, Wei and Ding, Yiwen and Hong, Boyang and Zhang, Ming and Wang, Junzhe and Jin, Senjie and Zhou, Enyu and others},
  journal={Science China Information Sciences},
  volume={68},
  number={2},
  pages={121101},
  year={2025}
}

@inproceedings{wei2022chain,
  title={Chain-of-Thought Prompting Elicits Reasoning in Large Language Models},
  author={Wei, Jason and Wang, Xuezhi and Schuurmans, Dale and Bosma, Maarten and Xia, Fei and Chi, Ed and Le, Quoc V and Zhou, Denny},
  booktitle={Advances in Neural Information Processing Systems (NeurIPS)},
  volume={35},
  pages={24824--24837},
  year={2022}
}

@inproceedings{wang2023selfconsistency,
  title={Self-Consistency Improves Chain of Thought Reasoning in Language Models},
  author={Wang, Xuezhi and Wei, Jason and Schuurmans, Dale and Le, Quoc and Chi, Ed and Narang, Sharan and Chowdhery, Aakanksha and Zhou, Denny},
  booktitle={International Conference on Learning Representations (ICLR)},
  year={2023}
}

@inproceedings{yao2024tree,
  title={Tree of Thoughts: Deliberate Problem Solving with Large Language Models},
  author={Yao, Shunyu and Yu, Dian and Zhao, Jeffrey and Shafran, Izhak and Griffiths, Thomas L and Cao, Yuan and Narasimhan, Karthik},
  booktitle={Advances in Neural Information Processing Systems (NeurIPS)},
  volume={36},
  year={2024}
}

@inproceedings{zhang2025cut,
  title={Cut the Crap: An Economical Communication Pipeline for LLM-based Multi-Agent Systems},
  author={Zhang, Guibin and Yue, Yanwei and Li, Zhixun and Yun, Sukwon and Wan, Guancheng and Wang, Kun and Cheng, Dawei and Yu, Jeffrey Xu and Chen, Tianlong},
  booktitle={International Conference on Learning Representations (ICLR)},
  year={2025},
  note={Reference for AgentPrune}
}

@article{wang2025agentdropout,
  title={AgentDropout: Dynamic Agent Elimination for Token-Efficient and High-Performance LLM-Based Multi-Agent Collaboration},
  author={Wang, Zhexuan and Wang, Yutong and Liu, Xuebo and Ding, Liang and Zhang, Miao and Liu, Jie and Zhang, Min},
  journal={arXiv preprint arXiv:2503.18891},
  year={2025}
}

@inproceedings{du2023improving,
  title={Improving Factuality and Reasoning in Language Models through Multiagent Debate},
  author={Du, Yilun and Li, Shuang and Torralba, Antonio and Tenenbaum, Joshua B and Mordatch, Igor},
  booktitle={International Conference on Machine Learning (ICML)},
  pages={8155--8168},
  year={2023}
}

@inproceedings{hendrycks2020measuring,
  title={Measuring Massive Multitask Language Understanding},
  author={Hendrycks, Dan and Burns, Collin and Basart, Steven and Zou, Andy and Mazeika, Mantas and Song, Dawn and Steinhardt, Jacob},
  booktitle={International Conference on Learning Representations (ICLR)},
  year={2021}
}

@article{cobbe2021training,
  title={Training Verifiers to Solve Math Word Problems},
  author={Cobbe, Karl and Kosaraju, Vineet and Bavarian, Mohammad and Chen, Mark and Jun, Heewoo and Kaiser, Lukasz and Plappert, Matthias and Tworek, Jerry and Hilton, Jacob and Nakano, Reiichiro and others},
  journal={arXiv preprint arXiv:2110.14168},
  year={2021}
}

@article{chen2021evaluating,
  title={Evaluating Large Language Models Trained on Code},
  author={Chen, Mark and Tworek, Jerry and Jun, Heewoo and Yuan, Qiming and Pinto, Henrique Ponde de Oliveira and Kaplan, Jared and Edwards, Harri and Burda, Yuri and Joseph, Nicholas and Brockman, Greg and others},
  journal={arXiv preprint arXiv:2107.03374},
  year={2021}
}

@article{wang2024survey,
  title={A survey on large language model based autonomous agents},
  author={Wang, Lei and Ma, Chen and Feng, Xueyang and Zhang, Zeyu and Yang, Hao and Zhang, Jingsen and Chen, Zhiyuan and Tang, Jiakai and Chen, Xu and Lin, Yankai and others},
  journal={Frontiers of Computer Science},
  volume={18},
  number={6},
  pages={186345},
  year={2024},
  publisher={Springer}
}

@book{sutton2018reinforcement,
  title={Reinforcement learning: An introduction},
  author={Sutton, Richard S and Barto, Andrew G},
  year={2018},
  publisher={MIT press}
}

@inproceedings{velickovic2017graph,
  title={Graph Attention Networks},
  author={Veli{\v{c}}kovi{\'c}, Petar and Cucurull, Guillem and Casanova, Arantxa and Romero, Adriana and Lio, Pietro and Bengio, Yoshua},
  booktitle={International Conference on Learning Representations (ICLR)},
  year={2018}
}

@inproceedings{wang2020minilm,
  title={MiniLM: Deep Self-Attention Distillation for Task-Agnostic Compression of Pre-Trained Transformers},
  author={Wang, Wenhui and Wei, Furu and Dong, Li and Bao, Hangbo and Yang, Nan and Zhou, Ming},
  booktitle={Advances in Neural Information Processing Systems (NeurIPS)},
  volume={33},
  pages={5776--5788},
  year={2020}
}

@inproceedings{roy2015solving,
  title={Solving General Arithmetic Word Problems},
  author={Roy, Subhro and Roth, Dan},
  booktitle={Proceedings of the 2015 Conference on Empirical Methods in Natural Language Processing (EMNLP)},
  pages={1743--1752},
  year={2015}
}

@inproceedings{patel2021nlp,
  title={Are NLP Models really able to solve simple math word problems?},
  author={Patel, Arkil and Bhattamishra, Satwik and Goyal, Navin},
  booktitle={Proceedings of the 2021 Conference of the North American Chapter of the Association for Computational Linguistics (NAACL)},
  pages={2080--2094},
  year={2021}
}

@inproceedings{ling2017program,
  title={Program Induction by Rationale Generation: Learning to Solve and Explain Algebraic Word Problems},
  author={Ling, Wang and Yogatama, Dani and Dyer, Chris and Blunsom, Phil},
  booktitle={Proceedings of the 55th Annual Meeting of the Association for Computational Linguistics (ACL)},
  pages={158--167},
  year={2017}
}

\end{document}